# *FedHIL*: Heterogeneity Resilient Federated Learning for Robust Indoor Localization with Mobile Devices

DANISH GUFRAN

Colorado State University, Fort Collins, CO, USA, Danish.Gufran@colostate.edu

SUDEEP PASRICHA

Colorado State University, Fort Collins, CO, USA, Sudeep@colostate.edu

Indoor localization plays a vital role in applications such as emergency response, warehouse management, and augmented reality experiences. By deploying machine learning (ML) based indoor localization frameworks on their mobile devices, users can localize themselves in a variety of indoor and subterranean environments. However, achieving accurate indoor localization can be challenging due to heterogeneity in the hardware and software stacks of mobile devices, which can result in inconsistent and inaccurate location estimates. Traditional ML models also heavily rely on initial training data, making them vulnerable to degradation in performance with dynamic changes across indoor environments. To address the challenges due to device heterogeneity and lack of adaptivity, we propose a novel embedded ML framework called *FedHIL*. Our framework combines indoor localization and federated learning (FL) to improve indoor localization accuracy in device-heterogeneous environments while also preserving user data privacy. *FedHIL* integrates a domain-specific selective weight adjustment approach to preserve the ML model's performance for indoor localization during FL, even in the presence of extremely noisy data. Experimental evaluations in diverse real-world indoor environments and with heterogeneous mobile devices show that *FedHIL* outperforms state-of-the-art FL and non-FL indoor localization frameworks. *FedHIL* is able to achieve 1.62× better localization accuracy on average than the best performing FL-based indoor localization framework from prior work.

## 1 INTRODUCTION

Indoor localization is the process of determining the location of a device or user within an indoor environment, such as a building or a warehouse [1]. This technology is becoming increasingly important in critical applications such as emergency response, security, and healthcare, where knowing the precise location of individuals or assets can be critical for saving lives and preventing harm [2]. Many efforts from industry and academia are attempting to build indoor localization solutions because of its promising future [3]. One foundational technology for indoor localization is Wi-Fi, which is ubiquitously deployed across indoor locales and can thus be cost-effectively used as part of indoor localization solutions [4].There are two main types of Wi-Fi-based indoor localization systems: propagation model-based and fingerprinting-based [5]-[8].

Propagation model-based methods estimate the distance between the transmitter (e.g., Wi-Fi access point (AP)) and receiver (e.g., mobile device carried by user) using propagation models such as the free space model. Then, by computing the relative locations of the receiver to specific transmitters, these methods estimate the user's coordinates in the indoor environment. Propagation model-based methods have advantages but also many limitations. These methods require prior knowledge of the AP locations, which may make them unsuitable for environments where AP locations are unknown or deliberately obscured due to security concerns [9]. Additionally, signal propagation in the indoor environment is dynamic and influenced by unpredictable factors such as obstacles and people, which can impact the accuracy of propagation model-based methods. Other factors, such as signal attenuation, multi-path fading, and shadowing, can also distort signals and result in inaccurate location estimates [10]. These limitations motivate the need for more robust localization methods.

Fingerprinting-based methods extract representative and distinguishable characteristics from indoor environments, called fingerprints [11]. In these methods, any radio frequency (RF) signal, such as Wi-Fi, Bluetooth, and Zigbee, can be used to estimate the user's location in indoor environments [12]. However, Wi-Fi is one of the most popular and widely used RF signals for indoor localization due to its availability in almost all indoor locales, without the need for deploying expensive custom APs [1]-[3]. Compared to other approaches, Wi-Fi fingerprinting can provide more accurate location estimates as it has a higher sampling rate and can enable capturing a larger amount of signal data. Additionally, Wi-Fi signals can penetrate through walls and other obstacles, providing more comprehensive coverage of indoor environments [13].

A Wi-Fi fingerprinting system typically consists of an offline and an online phase [1]-[4]. In the offline phase, Wi-Fi signal characteristics (e.g., received signal strength or RSS) at predefined reference points (RPs; representing specific locations within the indoor environment) are collected to build a fingerprint database. In the online phase, the system extracts a fingerprint from the observed Wi-Fi signals in the same way and matches the observed fingerprint to those saved in the database. The user's location is then estimated by determining the RP whose fingerprint is most similar to the observed fingerprint captured by the user's mobile device. While Wi-Fi fingerprinting has many advantages, there are also several limitations [9], [10]. One major limitation is device heterogeneity, which can result in variations in signal strength and accuracy [10], [13]. Different mobile devices can have varying Wi-Fi chips, antennas, and software stacks that can impact the captured Wi-Fi signal characteristics. This can lead to inconsistencies in the Wi-Fi fingerprints collected from different devices, making it difficult to accurately estimate the user's location. Additionally, changes in the Wi-Fi environment, such as interference from movement of people or changes in the configuration of the Wi-Fi APs can also impact the accuracy of the location estimates [13].

Machine learning (ML) models can be trained to learn features from Wi-Fi data (e.g., RSS) and use them to provide more accurate location predictions and address some of the limitations mentioned above [14]. One approach is to use supervised learning, where the system is trained using a dataset of known locations (RPs) and their corresponding Wi-Fi RSS fingerprints. In this approach, the ML model learns to map the input (Wi-Fi RSS fingerprint) to the output (known RP). The ML model structure can vary, e.g., they can be fully connected deep neural networks or convolutional neural networks [15]. Another approach is to use unsupervised learning, where the system is not provided with labeled data but instead tries to identify patterns and structure in the data [16]. In this approach, the ML model can be used to cluster similar Wi-Fi RSS fingerprints together, which can help to identify regions of similar signal strength in the indoor environment [17]. This information can be used to provide location estimates based on the nearest cluster or the centroid of the cluster.

Another approach to training ML models for indoor localization is in a distributed manner by using federated learning (FL). This approach allows multiple devices to collaboratively train a model without sharing their data, which is particularly important in scenarios where data privacy is a concern, such as in healthcare or industrial tracking [18]. In FL, the devices do not share their location data directly with a central server, but instead share the updated model parameters from a local model (LM) with a server, which aggregates the received parameters from each device to update a global model (GM) maintained at the server, while keeping the data on each device private [19]. FL can be used in scenarios where devices are heterogeneous, i.e., with different capabilities or hardware [20]. The FL approach allows for each device to contribute to the model training, regardless of its capabilities, providing a more comprehensive and diverse dataset. But the FL technique also has a major limitation: the performance of FL depends on the quality and diversity of the data contributed by each device [18]-[20]. If the data provided by different devices is noisy, it can lead to a degradation of the model performance, as the model may struggle to generalize across the data. This is because data from different devices may contain noise or outliers that fall outside the true distribution of the data. In such cases, effective feature selection of the trained model parameters before aggregating them becomes extremely important.

In this paper, we propose a novel embedded ML framework called *FedHIL*, that utilizes a neural network-based FL algorithm tailored for the indoor localization domain, to address the multiple challenges of device heterogeneity, model degradation, and privacy concerns. Our key contributions are as follows:

- We propose a novel and strategically tailored data augmentation strategy using unsupervised learning to address the variations in the Wi-Fi RSS fingerprints caused by device heterogeneity.

- We propose a combination of a lightweight neural network (supervised learning) within an FL setting that continually updates the underlying supervised model to maintain accuracy within the device heterogeneity-resilient and privacy-preserving framework.
- We propose a unique and strategic federated aggregation technique to contribute to the robustness of the proposed framework for indoor localization, thereby preventing degradation of the model's performance, caused by noisy or outlying data.
- We deploy the proposed *FedHIL* framework on multiple heterogeneous mobile devices and conduct evaluations across real-world building environments, and comprehensively compare its performance against several state-of-the-art FL- and non-FL based indoor localization frameworks.

## 2  RELATED WORK

In recent years, ML has become increasingly popular in Wi-Fi RSS fingerprinting based indoor localization [14]. Supervised learning methods such as K-nearest neighbor (KNN) [21], support vector machines (SVM) [22], random forest (RF) [23], and neural networks including deep neural networks (DNN) [24] and convolutional neural network (CNN) [25] have been widely used to improve localization accuracy. In particular, neural network-based approaches have shown promising results due to their ability to model complex non-linear relationships in the data [24], [25]. However, these supervised learning methods do not address the issues faced due to device heterogeneity. These methods also require a large amount of training data, which is problematic in real-world settings, where collecting large amounts of labeled fingerprint data may be too time consuming and costly, and thus impractical.

To overcome the limitations of supervised learning-based approaches, some prior works employ unsupervised learning-based approaches such as principal component analysis (PCA) [26], k-means clustering [27], and non-negative matrix factorization (NMF) [28]. These approaches aim to discover the underlying structure in the Wi-Fi RSS fingerprint data without the need for labeled data (RP of the corresponding fingerprints). Such methods can be useful in cases where obtaining labeled training data is difficult or costly. The works in [26]-[28] utilize various unsupervised learning methods to extract the most important features from the data and represent it in a lower-dimensional space. This can help to improve the performance of the ML model by reducing the complexity of the data and removing irrelevant or redundant features. While unsupervised methods can be useful for discovering underlying structure in the data, they are rarely as accurate as supervised methods in predicting location. Therefore, recent works propose a combination of supervised and unsupervised learning to create a more robust model to address device heterogeneity.

The hybrid approaches that combine supervised and unsupervised learning-based models take advantage of the strengths of the two unique learning approaches to improve the accuracy and robustness of indoor localization. The unsupervised learning methods are typically used to augment the training data, thereby diversifying the training data [29]. Augmenting the data can be beneficial because it can result in physically capturing lesser data for training while maintaining the diversity in the data. Additionally, augmenting the data can help reduce the effects of heterogeneity by generating more training examples with variations in the RSS. This improves the robustness of the model and its ability to handle RSS variations caused by device heterogeneity. The works in [30]-[33] make use of hybrid approaches. They concatenate the output from the unsupervised learning models and send it to a supervised learning model to yield stable localization results. This approach allows the supervised learning model to learn from the augmented data and extract the most relevant features from the data.

In recent years, many other frameworks have been proposed to address the challenge of device heterogeneity during indoor localization [34]-[38]. These frameworks employ a variety of techniques to address device heterogeneity, including multi-headed attention neural networks with a lightweight augmentation strategy [34][35], a combination of KNN and DNN [36], vision transformer neural networks with specialized augmentation strategies [37][38], and stacked autoencoders with a Gaussian process classifier (GPC) [39]. While these frameworks have shown promising results, they heavily rely on the initial training phase of the model, and any noisy or outlier data during the online phase can negatively impact the accuracy of the localization prediction.

A few recent works explore the use of federated learning (FL) to achieve collaborative learning for indoor localization [40]-[42]. As discussed earlier, FL allows multiple participating client devices to train an underlying GM at a server by sharing parameters from LMs on these devices that are trained with locally collected data. This approach preserves privacy during distributed learning as devices do not need to share their local (fingerprint) data with the server. The work in [40] proposes a centralized indoor localization method using pseudo-labels (CRNP). The authors use a stacked autoencoder for data augmentation and a DNN-based classifier with FL. During the online phase, the pseudo-labels generated from the DNN are used to retrain the model, and the retrained parameters are aggregated using the federated averaging (FedAvg) [43] method, which is a popular method for aggregation in FL. FedAeDNN [41] uses an autoencoder and DNN with FedAvg, while [42] proposes FedLoc, which uses a multi-layer perceptron and a federated stochastic gradient descent (FedSGD) [44] aggregation strategy. FedSGD is similar to FedAvg but uses a stochastic gradient descent (SGD) approach for model aggregation.

The ability to learn from a greater amount of data (from multiple clients while preserving their data privacy) has allowed these FL-based approaches to achieve promising results. However, these methods are sensitive to noisy or outlier data, which can skew the model's performance and lead to poor generalization and overfitting. To overcome these issues, in this work we propose a novel FL-based framework called *FedHIL* that builds on the merits of the FL-based methods discussed earlier. *FedHIL* utilizes a novel and uniquely tailored unsupervised learning-based ML model for efficient data augmentation, and the augmented data is further trained with an underlying lightweight supervised model that is also robust to device heterogeneity. This proposed combination is used in a distributed FL environment, where multiple client devices are used to update the model with their new and heterogeneous data. As FL can cause the accuracy of the underlying ML model to degrade, we further propose a new domain-specific selective weight adjustment approach to preserve the ML model's performance for indoor localization, even in the presence of extremely noisy data.

## 3  HETEROGENEITY ANALYSIS OF WI-FI RSS FINGERPRINTS

Wi-Fi RSS is a measurement of the power level of the wireless signal received from a Wi-Fi AP by a Wi-Fi receiver, such as those found in all smartphones, tablets, and laptops today. The RSS value is expressed in units of decibels (dB) relative to a milliwatt and represents the strength of the Wi-Fi signal received by the receiver. These values typically range from -100dB to 0dB [45]. A value of -100dB indicates that no signal is received, while a value of 0dB represents the strongest signal strength that can be received. However, it is rare to capture an RSS value of 0dB in real-world environments, as the signal strength is typically affected by factors such as signal attenuation, reflection, and multipath propagation. In practice, the RSS values captured during Wi-Fi fingerprinting are often in the range of -80dB to -30dB, with values below -70dB generally considered weak and values above -50dB generally considered strong [45]. The exact threshold values for strong and weak RSS signals may vary depending on the specific Wi-Fi chipset and the environment in which it is used.

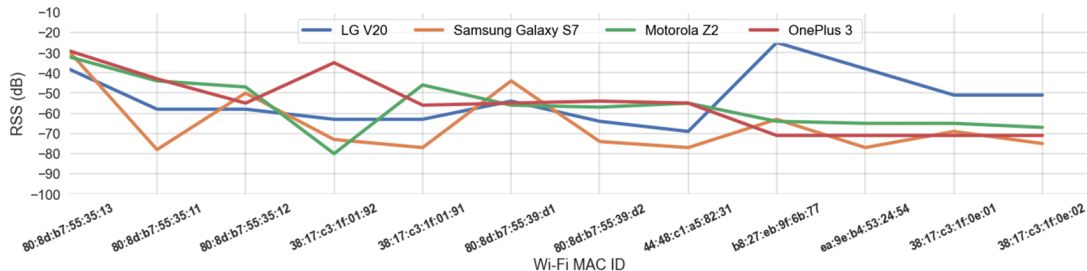

Figure 1: RSS fingerprints of visible Wi-Fi APs collected using four different smartphones at the same indoor location.

The observed Wi-Fi RSS values at any indoor location are impacted by many factors, such as the distance between the receiver and the AP, the type of obstacles and interference in the signal path, the signal propagation characteristics of the wireless channel, and most importantly, device heterogeneity. Device heterogeneity is a major problem in indoor localization because it significantly affects the accuracy and reliability of localization results. This is because indoor localization systems rely on Wi-Fi RSS fingerprints to estimate the location of the target device, and the RSS values captured by different devices can vary significantly due to the differences in their hardware, software, and antenna configurations [13]. Devices with the same Wi-Fi chipset can exhibit different Wi-Fi RSS readings due to heterogeneity in their firmware stacks. Variations in firmware configurations, calibration settings, and signal processing algorithms can lead to discrepancies in the RSS values reported by different devices, even if they share the same underlying Wi-Fi chipset. Since the RSS values captured by different devices at the same location can vary, the data used to train ML models that are part of indoor localization frameworks may not accurately represent the precise RSS values incident at a location. This can result in poor generalization of the model, leading to inaccurate localization results. To better understand device heterogeneity-based variations, we show results from a study we conducted that captured Wi-Fi RSS values at a specific indoor location in a real-world building setting with four different smartphones: LG V20, Samsung Galaxy S7, Motorola Z2, and OnePlus 3, as shown in Figure 1. The x-axis represents the MAC IDs (Media Access Control address) of the Wi-Fi APs, and the y-axis shows the RSS values captured by each smartphone. Ideally, the RSS value from each AP should be the same across the smartphones. From Figure 1, we observe that the values can differ significantly. For instance, for the AP with the MAC address '80:8d:b7:55:35:11', the LG V20 recorded an RSS value of -58 dB, while the Samsung Galaxy S7 recorded -78 dB. Similarly, for the AP with the MAC address '38:17:c3:1f:01:92', the Motorola Z2 and Samsung Galaxy S7 recorded -80 dB and -73 dB respectively, while LG V20 and OnePlus 3 recorded -63 dB and -35 dB respectively. We comprehensively analyzed such RSS variations across different buildings and a larger pool of heterogeneous smartphones. We utilized this information to both design and validate our *FedHIL* framework. Our data augmentation and FL strategies have been specifically designed to improve the accuracy and robustness of the indoor localization system by mitigating the impact of RSS variations.

## 4   BACKGROUND ON FEDERATED LEARNING (FL) FOR INDOOR LOCALIZATION

FL is a powerful technique that enables training of ML models across multiple decentralized client devices. In an FL system, training data is kept on the local devices (decentralized client devices), and the ML model is trained on each device. The updated model parameters are then transmitted and aggregated to create a global model that reflects the collective knowledge of all the devices. This approach provides many benefits, including enhanced data privacy, improved scalability, and long-term reliability of the model. In the context of indoor localization with Wi-Fi RSS data, FL can be used to train models that accurately predict the location of a device based on its local Wi-Fi RSS readings. Figure 2 shows an overview of how FL can be utilized for indoor localization. In this scenario, a centralized server is used to maintain a global model (GM), which is pretrained with a subset of the Wi-Fi RSS data. The GM represents the collective knowledge of all the local client devices participating in the FL process. Initially, this GM is first sent to client devices (e.g., via a localization app) which use it to make predictions based on their local Wi-Fi RSS readings. After receiving the GM from the centralized server, a client device can retrain the GM with its own local RSS data to create a local model (LM). The local training process involves retraining the received GM from the server using the client device's own RSS fingerprint dataset. As the client device's local RSS data is used to update the weights (model parameters) of the LM, it helps to improve the accuracy of the GM by accounting for local variations in the Wi-Fi RSS data. To collect the local Wi-Fi RSS readings, a client device at a specific indoor location captures Wi-Fi signals from nearby APs and measures their strength. The RSS values are combined into an RSS fingerprint, which is then sent to the LM to estimate the device (and user's) location. The client device can display the location on the user's device on a map of the building floorplan and this can be used for a variety of applications such as indoor navigation, location-based advertising, and asset tracking. The trained parameters from the client device's LM are periodically sent back to the centralized server, where they are aggregated with the original GM and parameters received from other devices, to update the GM, which is then sent to all client devices (Figure 2). This process of updating the GM with the contributions from multiple client devices helps to achieve better accuracy and scalability than traditional ML approaches.

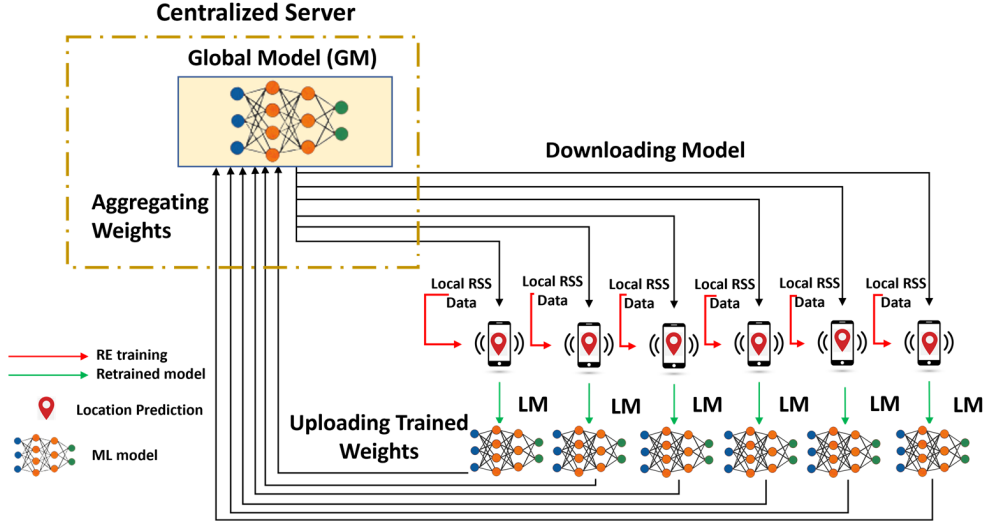

Figure 2: Federated learning (FL) overview using multiple client devices for indoor localization.

While there are several popular aggregation techniques used in FL, state-of-the-art approaches that use FL for indoor localization ([40]-[42], discussed earlier in Section 3) employ FedAvg [43] and FedSGD [44]. These aggregation techniques offer a simple and efficient way to combine models from different clients. We describe these aggregation approaches next. In Section 5, we describe our proposed FL-based localization framework.

### 4.1 Federated Averaging (FedAvg)

The FedAvg algorithm, first introduced by [43] in 2016, aims to train a GM using data available on multiple devices without transferring the data to a central server. The algorithm can be broken down into six steps: 1) initialization of a GM on a central server, 2) selection of a subset of clients to participate in each round of training, 3) sending of the GM to the selected client devices, 4) local training of the model on each device using an SGD algorithm, 5) sending of the updated models back to the central server, and 6) aggregation of the models using a weighted average. The aggregation strategy used in the FedAvg technique is shown in the equations below.

$$W_{client} = W_T - \eta * \nabla lf(W_T) \qquad (1)$$

$$W_{global} = \sum_{i=1}^{N} \frac{K_i + W_{client\,(i)}}{K} \qquad (2)$$

where $W_{client}$ are the weights generated from the LM using SGD, $W_T$ are the weights from the GM at the beginning of the round, η is the learning rate, and $\nabla lf(W_T)$ is the gradient of the loss function with respect to the GM weights. The $W_{client}$ obtained from (1) is used to update the GM using the weighted average method, as shown in (2). $W_{global}$ are the updated GM weights, N is the number of client devices, $K_i$ is the number of local data samples of the $i^{th}$ client, and K is the total number of local data samples in all clients. (2) calculates the weighted average of the LMs based on the number of local data samples available at each client. Clients with more data contribute more to the GM, while clients with less data have a smaller impact on the final model.

FedAvg is a popular technique for training ML models in distributed settings. However, it suffers from limitations such as overfitting and low accuracy over time in the presence of high noise in environments. For applications such as indoor localization using Wi-Fi RSS data, the data can be preprocessed and cleaned on each client device, and local ML models can be trained using an SGD algorithm. The resulting models can be aggregated using the FedAvg algorithm to predict the location within an indoor environment without the need for centralized data storage or processing. Overall, FedAvg is a powerful technique that can be applied to a wide range of applications, including those involving distributed data.

## 4.2 Federated Stochastic Gradient Descent (FedSGD)

FedSGD [44] was introduced as another distributed ML technique, similar to FedAvg. The main difference between FedSGD and FedAvg is the aggregation strategy, where FedAvg uses a weighted average of the LMs, while FedSGD uses a simple averaging approach. Like FedAvg, FedSGD also follows the same steps, except for the aggregation step. In FedSGD, gradients are calculated locally on individual devices using available data, which are sent to a central server for aggregation and to update the GM. Gradients represent the direction and magnitude of change required to minimize the loss function, which is used to update the weights in each iteration of the training process. The equations below capture the aggregation strategy used in FedSGD.

$$G_{client} = \nabla lf(W_T) \quad (3)$$

$$G_{avg} = \frac{1}{N}\sum_{i=1}^{N} G_{client\,(i)} \quad (4)$$

$$W_{global} = W_T - \eta * G_{avg} \quad (5)$$

where $G_{client}$ are the gradients of the client models' loss functions with respect to the GM's weights, which is represented as $W_T$. These gradients are sent to the server for aggregation. The $G_{avg}$ are the averaged gradients of all the client devices. *N* is the number of client devices and $G_{client\,(i)}$ are the gradients computed using (3) per client. $W_{global}$ are the new GM weights obtained by subtracting the product of the learning rate ($\eta$) and the average of the client gradients ($G_{avg}$) obtained from (4). In this way, FedSGD technique uses the client gradients instead of the client weights (as done in FedAvg) to compute the average.

Much like FedAvg, FedSGD is beneficial in training ML models on decentralized and heterogeneous data sources, while maintaining privacy. In indoor localization based on Wi-Fi RSS data, FedSGD can be suitable for scenarios with a small number of clients and less imbalanced (or noisy) data. It can also be useful when communication bandwidth is limited, as its communication cost is lower than FedAvg.

## 5 *FEDHIL* FRAMEWORK

In this section, we describe our proposed *FedHIL* framework for FL-based indoor localization. The framework follows the general principles of FL, as discussed earlier. However, it also includes modifications that have been tailored to address some of the key challenges associated with device heterogeneity and long-term robustness. The first modification is in the GM architecture, which allows the framework to aggregate data from different sources and train a model that can generalize well across different devices. The second modification is in the LM architecture, which enables each device to train its model on local data while ensuring that the data remains private and secure. Finally, the third modification is in the federated aggregation approach, which ensures that the GM is updated with new information from different LMs via communication between the LM and GM. By integrating these modifications into the FL framework, *FedHIL* aims to overcome some of the limitations of traditional FL methods for the indoor localization problem. In the following sections, we describe each of these mechanisms, providing a comprehensive overview of the *FedHIL* framework and its capabilities.

### 5.1 Global model (GM) architecture

The GM architecture plays a critical role in achieving high accuracy and scalability. The GM architecture consists of two phases, namely, the offline training and online training phase, as shown in Figure 3.

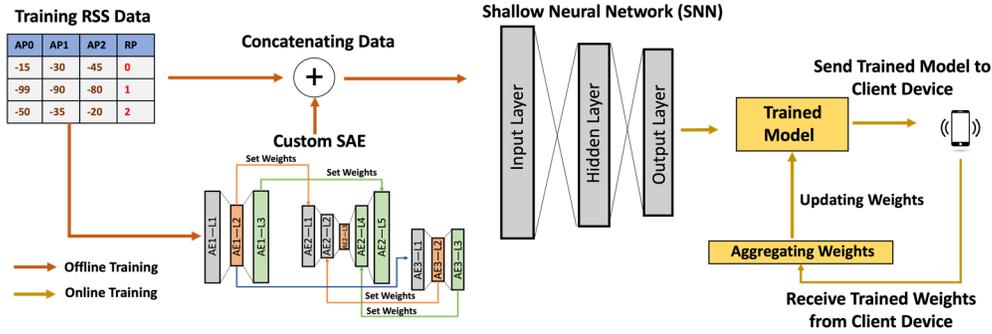

Figure 3: GM architecture for *FedHIL*

The offline training phase consists of two components: A data augmentation strategy using a custom stacked autoencoder (SAE) and a shallow neural network (SNN) model. Data augmentation using a custom SAE is a key component of the proposed framework. The custom SAE learns to extract meaningful features from the input RSS data and generates augmented data that can help improve the accuracy of the model. The output from the custom SAE is concatenated with the original RSS data and then sent for training the SNN model. The SNN model is chosen to have a lightweight architecture, which makes it well suited for mobile devices. Once the SNN model is trained, it is made available as the GM. The offline training phase is complete at this point, and the trained GM is then made available for deployment on client for prediction.

In the online training phase, the client device makes a copy of the GM and stores it locally. The client device uses the trained GM, now called the LM, to make predictions for the user with an unknown location. While this prediction is happening, the client device also retrains the LM using its local data. The updated trained weights in the LM are then sent back to the server, which holds the GM. The server receives the updated weights only and aggregates them with the GM, thus updating the GM without looking at the raw RSS data from the client devices. This process enables the GM to adapt to new heterogeneous data from multiple clients while maintaining the privacy of the client devices.

### 5.1.1 Custom Stacked Autoencoder (SAE) neural network

To enhance the collected RSS data prior to training, we customize an SAE neural network model to create augmented RSS data. Our SAE model comprises of multiple autoencoders (AE) that are trained one layer at a time using a layer-by-layer training approach. The AE neural network consists of three key components: an Encoder, a Decoder, and a Bottleneck layer. Each of these components is comprised of one or more fully connected layers with varying numbers of neurons. The encoder layers are responsible for compressing the input data and reducing its dimensionality, while the decoder layers are utilized to reconstruct a noisy representation of the input data back to its original dimension. The bottleneck layer represents the input data at its lowest dimension and reflects the most salient and significant features in the data.

The encoder layers consist of fully connected layers with decreasing numbers of neurons in each layer. These layers perform the compression and dimensionality reduction operations. By reducing the number of neurons in each layer, the encoder layers effectively reduce the dimensionality of the input data. Non-linear activation functions, such as ReLU (Rectified Linear Unit), are used in these layers to introduce non-linearity and capture complex relationships in the data. The bottleneck layer is positioned at the center of the AE and has the lowest dimensionality among all the layers in the network. It acts as the bottleneck for information flow, representing the most salient and significant features of the input data. The bottleneck layer typically has the least number of neurons, allowing it to capture the essential information while discarding less relevant or redundant information. The reduced dimensionality in this layer helps in extracting the most discriminative features from the input data. The encoder can be considered a neural network that transforms high-dimensional input data into lower-dimensional data (at the bottleneck layer).

The decoder layers mirror the structure of the encoder layers but in reverse order. They consist of fully connected layers with increasing numbers of neurons in each layer. These layers are responsible for reconstructing the compressed representation back to the original dimension. By increasing the number of neurons in each layer, the decoder layers expand the compressed representation and attempt to reconstruct the input data. The decoder layers are designed to reconstruct the data in the same way it was reduced, ensuring the preservation of relevant information and the underlying structure of the data. The layer-by-layer training approach employed in the SAE model is specifically designed to ensure minimal noise during the reconstruction process in the decoder. This approach allows each layer to learn an effective representation of the data before proceeding to the next layer. By gradually building up the complexity of the reconstruction, the network can better capture important details and relationships within the data. This results in more stable training and better overall performance of the model.

The proposed SAE model involves stacking three AEs together, each having its own fully connected layers, as shown in Figure 3. AE1 and AE3 are trained layer-wise, with the output of the second layer (L2) of AE1 being fed as input to AE3. AE2 is positioned as illustrated in Figure 3 and receives the trained weights from AE1 and AE3. The SAE model is trained layer-by-layer to minimize the reconstruction error between the input and output data. The advantage of this approach is that it allows the network to learn an effective representation of the input data at each layer before proceeding to the next one, rather than attempting to train all the layers simultaneously. This can result in a more stable training process and better overall performance. The SAE model is leveraged for RSS data augmentation by training it on the training RSS data created during the offline phase. The SAE can subsequently generate synthetic RSS measurements that can be used to augment the training dataset. The synthetic RSS data emulates the variations that can occur across different devices, potentially improving the system's robustness and generalization by providing a more diverse set of training data that accurately represents variations across different devices.

*5.1.2 Shallow Neural Network (SNN)*

The SNN model is an artificial neural network that receives concatenated input from both the training and augmented RSS data. Our SNN model is comprised of three fully connected layers, namely, the input layer, the hidden layer, and the output layer. The input layer is responsible for receiving the concatenated input data, which is then passed on to the hidden layer. The hidden layer is where the complex processing of the input data occurs, using various activation functions, such as rectified linear unit (Relu), and weights assigned to each neuron in the layer. The output of the hidden layer is then passed on to the output layer, which provides the final result of the model's prediction. One advantage of our SNN model is that it is lightweight (it has only a single hidden layer), making it easy to deploy on client devices. This is particularly beneficial for indoor localization as we require real-time prediction and processing of data, and a lightweight model minimizes the computational resources required. Furthermore, the augmented data used in the training of the SNN model allows it to generalize for device heterogeneity during indoor localization. This means that the model is able to recognize and predict the location of a user, even if they are using different devices to access the application.

## 5.2 Local model (LM) architecture

The client device receives the trained GM from the central server, as illustrated in Figure 4. Once received, the client device clones the trained GM and stores it locally as the LM, which is used to make predictions.

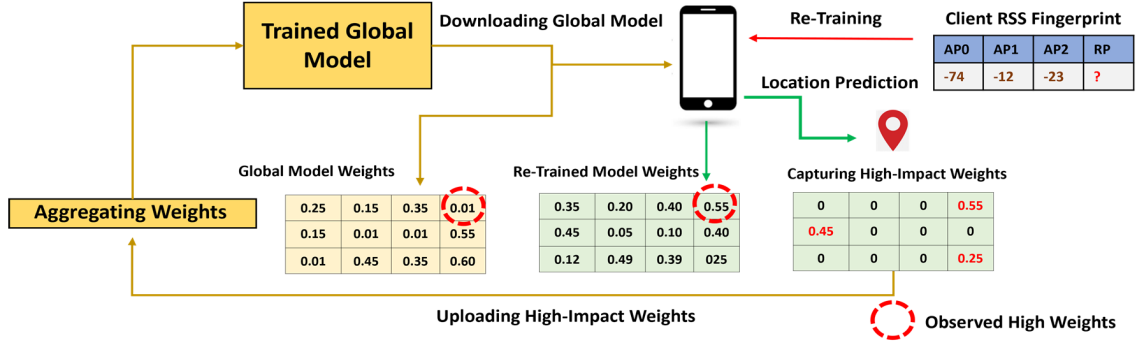

Figure 4: LM architecture for *FedHIL*

To enhance the accuracy of predictions, the client device scans for nearby Wi-Fi APs and stores their RSS data locally. This stored RSS data is used to inference with the LM and to retrain the LM to better fit the user's environment. This local training is supervised, and the client device uses its stored data to update its own local model through backpropagation with stochastic gradient descent. Gradients are calculated based on the loss function of the local model. The LM is retrained over a small number of epochs to ensure that the process has low overhead on the resource-constrained client device. After the LM is retrained, the client device uses an absolute distance method, which is part of the *FedHIL* aggregation technique (discussed in Section 5.3), to select only a subset of LM weights to upload to the server for aggregation. This is done to protect the GM from noisy, low impacting weights. By only transmitting a high impact subset of weights, not only are the communication requirements for the FL process relaxed, but overall accuracy can also be improved. Throughout this process, multiple client devices participate in the prediction and contribute to updating the GM with new heterogeneous device-specific RSS data. Figure 4 illustrates the selection process of the high-impact weights using the absolute distance method, which is discussed in more detail in the following subsection.

### 5.3 *FedHIL* aggregation technique

We now describe our FL aggregation method that has been specifically designed to address the challenges of heterogeneity and long-term accuracy robustness in indoor localization. Traditional FL techniques may cause the ML model to degrade in accuracy over multiple rounds of client-side retraining, due to noisy data collected from the client devices. These outliers in the online training phase can skew the ML model's performance and cause it to degrade. Our *FedHIL* aggregation technique addresses this issue by identifying selected features in the trained LM of the client during the online training phase. The selected features are the high-impacting features in the locally collected RSS fingerprints of the client device, calculated by finding the absolute distance of the weights between the trained LM and the cloned GM that each client maintains before retraining it. This absolute distance represents the changes in the weights for the visible features in the LM. Our technique then filters these absolute weights by only capturing the top H% of the highest changes in the weights. This means that the top H% of the largest magnitude difference between the LM and GM are captured, and the background noise and potentially noisy distribution in the weights picked up by the client device are filtered out. The top H% of weights from the client are then sent back to the server for aggregation. The received weights from the client device are then averaged with the existing GM in the server via simple averaging. This process is repeated with participating client devices, with each device contributing to the improvement of the ML model's performance. The equations below show a mathematical representation of the proposed technique:

$$W_{client} = W_T - \eta * \nabla\, lf(W_T) \qquad (6)$$

This equation is the same as in FedAvg, where $W_{client}$ are the weights generated from the LM using SGD, $W_T$ are the weights from the GM at the beginning of the round, $\eta$ is the learning rate, and $\nabla\, lf(W_T)$ is the gradient of the loss function with respect to the GM weights.

$$W_{abs} = |W_{global} - W_{client}| \qquad (7)$$

$$I_{High} = \psi(W_{abs}, H) \tag{8}$$

Equations (7), and (8) identify the high-impact weights in the $W_{client}$ from equation (6). Equation (7) involves calculating the absolute difference between the weights in the GM ($W_{global}$) and the weights of the LM ($W_{client}$). By calculating the absolute difference, we can identify the areas where the GM and LM differ the most. Equation (8) involves determining the indices of the top *H*% of weights. The variable *H*% represents the top percentage of the largest magnitude difference between the GM and the LM. The $\psi$ function is used in selecting the indices associated with the top *H*% weights in $W_{abs}$.

$$W_{High} = \begin{cases} W_{client}, & if\ I_{client} \in I_{High}, \\ 0, & otherwise \end{cases} \tag{9}$$

In (9), we capture the high impact weights in $W_{client}$ by setting the non-high-impact weights to 0 and keeping the high-impact weights the same as before. This is done to reduce the impact of client-to-client variation on the high-impact weights.

$$W_{new\ global} = \frac{1}{N}\sum_{i=1}^{N} W_{high} + W_{global} \tag{10}$$

Finally, (10) computes the new GM weights by taking the average of the selected high impact client model weights ($W_{high}$) and the original GM weights ($W_{global}$), where $N$ is the number of clients.

By following this technique, we update the GM with only high-impacting weights from the client devices. The proposed technique has several advantages. First, the high-impacting weights can prevent the ML model from overfitting, thereby preventing the GM model from degrading. Second, by only capturing the top H% of the weights, the proposed technique filters out noise, leading to better performance of the ML model. Third, by having multiple clients with heterogeneous data participating in this process, the overall system can achieve heterogeneity resilience. Lastly, since only a subset of the trained weights is shared, compression techniques can compress sparse (zero) weights to reduce the wireless communication volume and latency from the client devices to the server. These factors make *FedHIL* a promising FL approach for indoor localization.

## 6 EXPERIMENTAL RESULTS

### 6.1 Experimental Setup

In this section, we present the experimental setup for the evaluation of the proposed *FedHIL* framework. The aim of our experiments is to evaluate the performance of the *FedHIL* framework in real-world settings and compare it with other FL and non-FL based indoor localization frameworks.

Table 1: Trainable parameters in each layer in the custom SAE.

| AE1-L1 | AE1-L2 | AE1-L3 | AE2-L1 | AE2-L2 | AE2-L3 | AE2-L4 | AE2-L5 | AE3-L1 | AE3-L2 | AE3-L3 | Total |
|---|---|---|---|---|---|---|---|---|---|---|---|
| 14878 | 10414 | 14964 | 10414 | 5103 | 3570 | 7388 | 14964 | 7289 | 5103 | 7388 | 101475 |

In the *FedHIL* framework, we pre-train the GM using RSS data collected from a single smartphone. This is done to ensure that the framework is resilient towards heterogeneity and can be evaluated on any number of smartphones. To pre-train the GM, we first augment the original RSS data using a custom SAE, with the trainable parameters shown in Table 1. The custom SAE uses SGD for optimization with the MSE (mean square error) loss function. The custom SAE is trained for 700 epochs per layer, to generate the best results for the given data. We then concatenate the original and augmented data and use it to train the SNN model. The proposed SNN model in the *FedHIL* framework has a Relu non-linearity with 256 neurons in the hidden layer and uses the sigmoid function in the output layer for prediction. The SNN also uses the SGD optimization with the Sparse Categorical Crossentropy loss function and trained for 1200 epochs. The SNN model is trained with the trainable parameters shown in Table 2.

Table 2: Trainable parameters in each layer in the SNN model.

| Input Layer | Hidden Layer | Output Layer | Total |
|---|---|---|---|
| 22144 | 33024 | 15677 | 70845 |

After pre-training the GM, we make it available to each client device for testing and retraining. The client devices retrain the SNN model over 10 epochs during the online training phase, to fit the resource constraints in the client devices and report only the top H% of the high impacting features in the local model back to the server. We set the H parameter in the proposed *FedHIL* aggregation technique based on our sensitivity analysis presented in Section 6.2. To evaluate the performance of the *FedHIL* framework, we measured the localization accuracy using the Euclidean distance formula:

$$Euclidean\ Distance = \sqrt[2]{(X_2 - X_1)^2 + (Y_2 - Y_1)^2 + (Z_2 - Z_1)^2} \quad (11)$$

where $(X_1, Y_1, Z_1)$ represents the ground truth location coordinates and $(X_2, Y_2, Z_2)$ represents the predicted location coordinates. The accuracy is thus calculated based on the distance between the ground truth location and the estimated location of a device, which is based on the RSS data collected by the device. This approach allows us to assess the accuracy of the framework across multiple devices and scenarios, ensuring that the framework can be reliably used in real-world settings.

We compared the performance of our *FedHIL* framework with three FL-based frameworks (CRNP [40], FedAeDNN [41], and FedLoc [42]) and four non-FL-based frameworks (ANVIL [33], SHERPA [36], VITAL [37] and WiDeep [39]). The FL-based frameworks were selected because they have shown promising results in addressing device heterogeneity, which is a key challenge. On the other hand, we chose the non-FL-based frameworks to provide a baseline for comparison and to highlight the improvements achieved using FL.

Table 3: Details of smartphones used in our experiments.

| Manufacturer | Model | Abbreviation | Wi-Fi Chipset |
|---|---|---|---|
| BLU | Vivo 8 | BLU | MediaTek Helio P10 |
| HTC | U11 | HTC | Qualcomm Snapdragon 835 |
| Samsung | Galaxy S7 | S7 | Qualcomm Snapdragon 820 |
| LG | V20 | LG | Qualcomm Snapdragon 820 |
| Motorola | Z2 | MOTO | Qualcomm Snapdragon 835 |
| Oneplus | Oneplus 3 | OP3 | Qualcomm Snapdragon 820 |

To collect data for our indoor localization experiments, we utilized six smartphones from different manufacturers, each equipped with a different Wi-Fi chipset, as shown in Table 3. We normalized the RSS values between 0 and 1 to train the ML model on a consistent and uniform dataset, which can lead to better performance and more accurate indoor localization.

We select five different buildings (Building 1-5) with different salient features. and performed experiments on paths of different shapes and with lengths ranging from 60 to 88 meters. Building 1 is made of wood, and concrete, and is surrounded by computers with large open areas. A total of 78 unique Wi-Fi APs were visible in this building. Building 2 and Building 3 are relatively new buildings with a mix of metal and wooden structures, open areas, and bookshelves with heavy metallic equipment. We observed 218 and 112 unique APs, respectively in these buildings. Building 1 and Building 2 are located in close proximity to each other and are separated by a single wall. The floorplans in these two buildings were chosen to consider the impact of interference from different sources in the two buildings. The mix of different materials and equipment in the buildings creates a complex indoor environment, which, together with interference from Wi-Fi APs in the other building, poses a challenge for indoor localization systems. Building 4 has several small offices, with a mix of metal, wood, and glass structures. The building was densely covered by 156 APs. Building 5 has laboratory equipment with a considerable amount of electronic and mechanical equipment and 125 visible APs.

The presence of metal and electronics in the vicinity of the tested paths in the buildings lead to noisy Wi-Fi fingerprints that complicate indoor localization efforts. We captured Wi-Fi RSS fingerprints during regular working hours, without artificially manipulating human occupancy. By collecting Wi-Fi RSS fingerprints during the course of entire days, including regular working hours, we were able to capture various forms of interference in Wi-Fi signal propagation. Our data collection methodology allowed us to incorporate various sources of interference, including dynamic factors such as population density and human movement within the building. By capturing data in this authentic setting, our aim is to gain a comprehensive understanding of the challenges and complexities encountered in real-world indoor localization, thereby enhancing the generalizability of our results. For the offline phase, we used five fingerprints per RP to train the GM, and for the online phase, we used one fingerprint per RP. We selected a granularity of 1-meter separation between RPs for our experiments, which we consider sufficient for localizing humans.

Table 4: Inference time latency for the SNN model on different smartphones in milli-seconds (ms).

| BLU | HTC | LG | MOTO | OP3 | S7 | Average |
|---|---|---|---|---|---|---|
| 73.35 ms | 62.05 ms | 64.81 ms | 76.87 ms | 64.16 ms | 67.25 ms | 68.08 ms |

To assess the lightweight nature of the SNN model implemented on the smartphones listed in Table 3, we measured the inference time latencies for each device using the SNN model during the online phase. The inference speed is influenced by various smartphone resources, including processor speed, memory, and other hardware capabilities. Based on our measurements, the average inference time latency across all smartphones was found to be 68.08 ms, indicating the efficient and lightweight performance of the SNN model on these devices. In the following subsections, we present sensitivity analysis, comparison of the proposed *FedHIL* framework with state-of-the-art indoor localization frameworks, and analysis for skew and scalability.

## 6.2 Sensitivity analysis for *FedHIL*

### 6.2.1 Selection of H parameter

In this section, we analyze the H parameter value within *FedHIL* and evaluate the framework's performance on real-world data. The parameter determines percentage of the trained weight values in the LM to be aggregated with the GM. Figure 5 shows experimental results for selecting the H parameter, whose value is shown ranging from 10% to 100%. The results shown are for the Building 5 floorplan (highly noisy environment) with the MOTO device used for training and OP3 device used for testing in the online phase. From the figure it can be observed that setting H to 10% causes the model to overfit easily, leading to a significant degradation in localization error (y axis), particularly after RP 5, as only a tiny proportion of the trained weights in the client LM are used to update the GM. Additionally, the model is heavily affected by noise, resulting in an increase in localization errors. Setting H to 20% yields the best results, where the model is most resilient towards noisy RP locations. Although the model with H=20% shows a few instances of high errors, they are much smaller than those observed for other H values. The H values from 30% to 100% show similar results, demonstrating some resilience towards noise, but they are still heavily affected by it, resulting in higher localization errors compared to the 20% value. The 100% H value is where all the trained LM weights are averaged with the GM, which is the same as the FedAvg aggregation approach. The approach shows high localization errors and is heavily affected by noise.

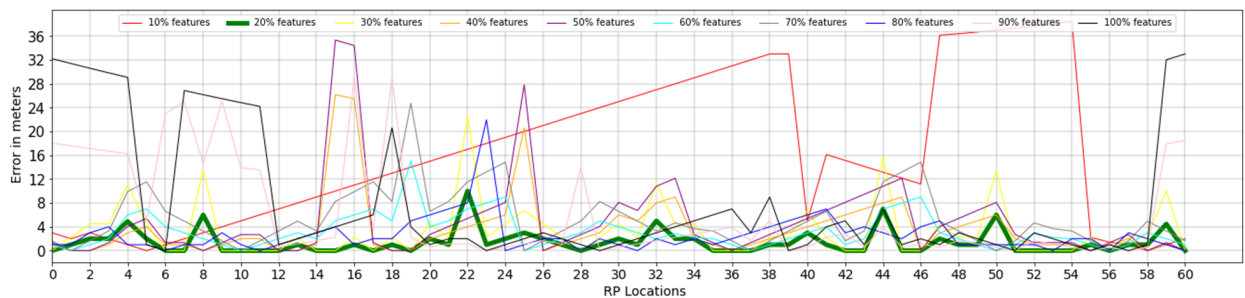

Figure 5: Sensitivity Analysis of RP Locations in Building 5 Floorplan for Selecting H Value in *FedHIL* Technique

To thoroughly analyze the impact of the H parameter value within the *FedHIL* framework and determine the optimal H value, we conducted experiments and plotted the average error for different H values. In Figure 6, we present the results, where each H value ranges from 10% to 100% with a step size of 10. Upon analyzing the plot, several observations can be made. Firstly, setting H to 10% resulted in the highest average error compared to other H values. This can be attributed to a significant loss of weights during aggregation, leading the model to easily overfit the data.

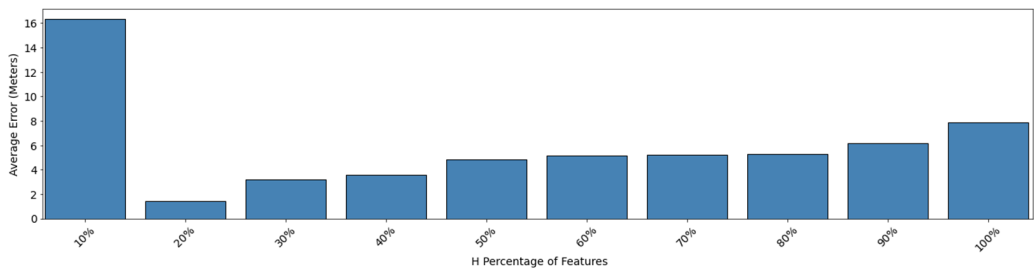

Figure 6: Average error in meters for different H value in *FedHIL* technique.

On the other hand, when we set H to 20%, we observed the lowest average error. This indicates that preserving 20% of the highest changing weights during aggregation strikes a balance between retaining important information and avoiding overfitting. As a result, the model achieves better generalization and performs more accurately. For H values of 30% and 40%, we observed slightly higher errors compared to H = 20%, but significantly lower errors than H = 10%. This suggests that these H values strike a trade-off between preserving important information and allowing for some level of noise or variability in the aggregated weights. Furthermore, H values of 50%,60%, 70%, 80%, and 90% showed similar average errors, albeit slightly higher than H = 30% and H = 40%. These values still maintain a good level of performance by preserving a considerable portion of the highest changing weights during aggregation. Lastly, when setting H to 100%, we observed the highest error among the range of 20% to 90%. Nevertheless, even H = 100% showed a significantly lower error compared to H = 10%, indicating that preserving all weights during aggregation still yields improvements over discarding all weights. Based on these findings, we conclude that setting H to 20% provides the optimal balance between weight preservation and error reduction, even in the presence of high noise in the floorplan (Building 5). We ultimately set the H parameter in *FedHIL* to 20%, as it also provided the best performance across other buildings. We used this value for *FedHIL* in all our subsequent experiments.

### 6.2.2 Client-Server Communication latency

In In this section, we analyze the communication latency between client devices and the server in our *FedHIL* framework. Figure 7 shows the average communication latency for different H values across all RPs in Building 5. We observe that as the H value increases, the communication latency also increases. This is because a higher H value involves transmitting more data from the client to the server, resulting in longer transfer times. However, the average communication latency differences between H values are relatively small, always remaining below 1 second. Analyzing and optimizing communication latency is crucial in FL as it directly impacts system performance and efficiency.

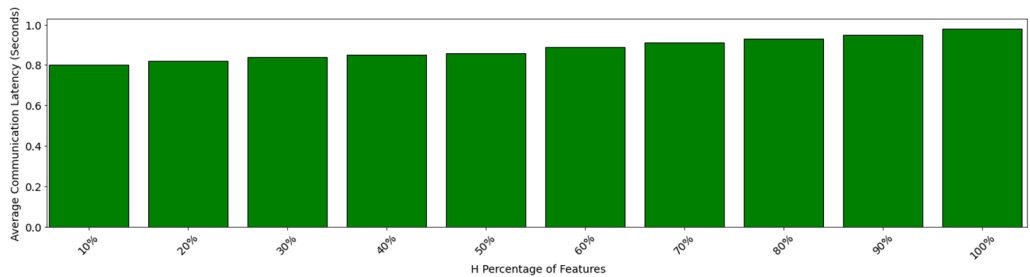

Figure 7: Average communication latency in seconds for different H value in *FedHIL* technique.

A higher communication latency can result in delays in aggregating the model updates, leading to slower convergence and overall training performance. On the other hand, minimizing communication latency enables faster synchronization, allowing the system to adapt and learn more quickly. Combining the findings from Figure 6, which illustrates the relationship between H values and average error, we observe that H = 20% achieves the lowest average error, indicating better localization accuracy. Additionally, H = 10% exhibits the lowest communication latency, but it compromises performance. Higher H values show slightly higher communication latency without significant improvements in accuracy. Considering these aspects, we conclude that H = 20% strikes a balance between performance and communication latency. It offers the optimal solution within the *FedHIL* framework, ensuring accurate localization while maintaining efficient communication.

### 6.2.3 Impact of custom SAE in FedHIL

In order to evaluate the impact of the custom SAE in addressing device heterogeneity challenges during indoor localization, we conducted a series of experiments comparing different configurations of the *FedHIL* framework. Specifically, we examined three situations during the offline training phase: *FedHIL* without SAE (no augmentation), *FedHIL* with a traditional SAE, and *FedHIL* with the custom SAE. To introduce data augmentation, we incorporated a traditional SAE into the *FedHIL* framework. The parameters of this traditional SAE were set to be identical to those of the custom SAE. However, it is important to note that the traditional SAE was not trained using a layer-by-layer approach, which is a key characteristic of our custom SAE. To assess the performance of the trained SNN model, we conducted experiments using MOTO as the training device. We compared the results obtained from these experiments with the three previously mentioned situations, as illustrated in Figure 8. This comparison allowed us to analyze the influence of the custom SAE on the accuracy and effectiveness of the *FedHIL* framework in addressing device heterogeneity challenges during the offline training phase.

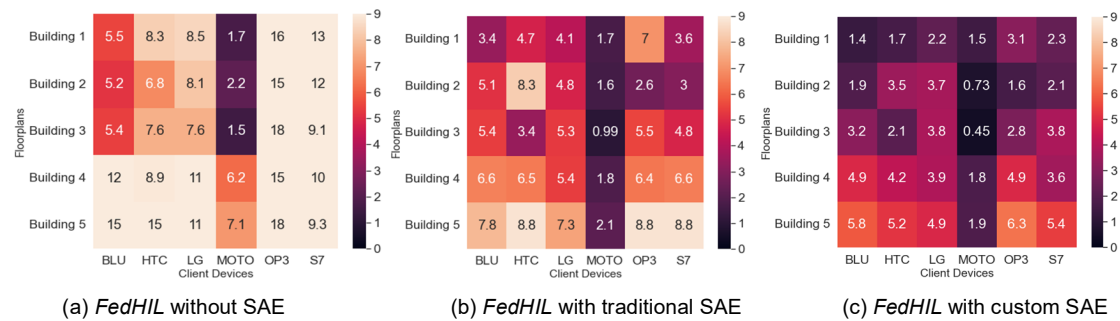

(a) *FedHIL* without SAE  (b) *FedHIL* with traditional SAE  (c) *FedHIL* with custom SAE

Figure 8: *FedHIL* with different SAE configurations, training device: MOTO (values in cells represent Euclidean distance).

Figure 8(a) illustrates the outcome of *FedHIL* without the SAE, Figure 8(b) depicts the results with the traditional SAE, and Figure 8(c) depicts the results with our proposed custom SAE. The evaluation aimed to assess the impact of the custom SAE on the localization performance across various client devices and building floorplans. To conduct the evaluation, we utilized the dataset consisting of six client devices and five different building floorplans, as discussed in section 6.1. For each combination of the client device and building floorplan, we measured the localization error using the Euclidean distance method. The localization error represents the discrepancy between the estimated position and the ground truth position of the client device. In Figure 8, each cell corresponds to a specific combination of client devices and building floorplan. The value displayed within each cell represents the average localization error observed across all available RPs within the corresponding building floorplan.

By analyzing these figures, we can clearly observe the importance of our proposed custom SAE in our *FedHIL* framework. The results indicate that the underlying SNN model without SAE only performs well in the absence of device heterogeneity. This can be observed from the lower localization errors for MOTO in the online phase, which is the same device that is used to train the model in the offline phase in Figure 8(a). Without SAE, the model fails to generalize on heterogeneous devices other than MOTO, including LG, OP3, and S7 (performance for BLU and HTC devices is slightly better, but still poor), and also in high noise environments, such as in Buildings 4 and 5. However, with the addition of the traditional SAE, we observe significant improvements in localization errors compared to those without using the SAE. The incorporation of the traditional SAE allows for some level of adaptation to device heterogeneity, leading to better performance across different devices and noise environments. This is evident from the reduced errors in the localization, especially for devices other than MOTO. However, it is important to note that the performance improvement is limited, and the model still struggles to achieve satisfactory accuracy in challenging scenarios, such as Building 4 and Building 5. On the other hand, with the custom SAE, we witness even more remarkable improvements in localization errors, surpassing the performance achieved with the traditional SAE. The custom SAE, trained in a layer-by-layer approach specifically tailored for our domain, effectively addresses the challenges posed by device heterogeneity. By augmenting the training data and diversifying its distribution, the custom SAE enables the SNN model to adapt and generalize across a wide range of devices and noisy environments.

The substantial reduction in localization errors with the custom SAE demonstrates its superior ability to enhance the robustness and accuracy of the *FedHIL* framework. The model now achieves reliable and accurate indoor localization across various devices and noise levels, indicating the effectiveness of our approach in mitigating the impact of device heterogeneity.

To summarize the effectiveness of the custom SAE within the *FedHIL* framework, we present a bar plot to compare the average error in meters across the three previously mentioned scenarios in Figure 9. The experimental setup aligns with the conditions in Figure 8.

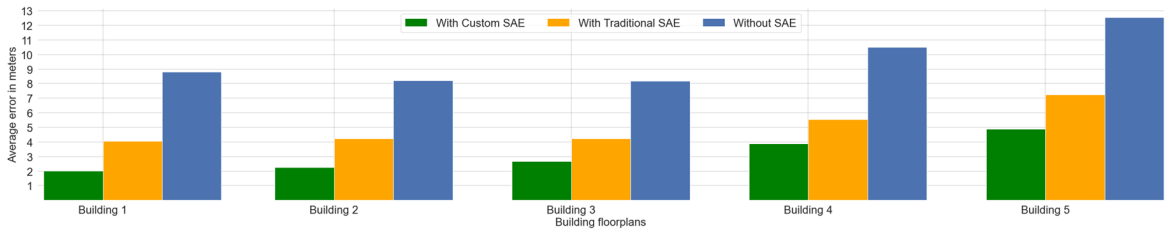

Figure 9: Impact of Custom SAE on Average Error in Building Floorplans

Figure 9 serves as compelling evidence showcasing the substantial positive impact of the custom SAE on the *FedHIL* framework's ability to handle device heterogeneity. The bar plot illustrates the average error per building floorplan, providing valuable insights into the framework's performance across different scenarios. The results are striking, revealing significant reductions in average errors when utilizing the custom SAE. Building 1 demonstrates a remarkable 77.1% reduction in errors compared to the scenario without SAE, and a noteworthy 50.2% error reduction compared to the traditional SAE. Building 2 follows closely, with a reduction of 72.59% compared to without SAE and a substantial 46.8% error reduction compared to the traditional SAE. Building 3 showcases a 67.19% error reduction compared to without SAE, and a commendable 36.4% error reduction compared to the traditional SAE. Building 4 exhibits a 63.08% reduction in errors compared to without SAE, and a notable 30.09% error reduction compared to the traditional SAE. Lastly, building 5 experiences a remarkable 60.9% reduction in errors compared to without SAE, along with a significant 32.36% error reduction compared to the traditional SAE. These findings powerfully emphasize the efficacy of the custom SAE in enhancing the framework's resilience to device heterogeneity, resulting in improved accuracy in indoor localization across diverse buildings. By effectively addressing the unique characteristics of each device, the integration of the custom SAE empowers the *FedHIL* framework to deliver more reliable and accurate localization results.

To provide a clear analysis of the impact of the custom SAE, we present a plot in Figure 10 that compares the original fingerprint with the augmented fingerprint generated using the custom SAE. Upon careful examination, we observe distinct variations in the augmented fingerprint compared to the original fingerprint. The custom SAE effectively captures and incorporates device heterogeneity, resulting in a more diverse and representative fingerprint. This augmentation process enables the model to learn and adapt to the unique characteristics of different devices, enhancing its ability to handle device heterogeneity. The differences between the original and augmented fingerprints are indicative of the custom SAE's capability to capture important features and patterns that can potentially arise due to device heterogeneity. These additional features contribute to a more comprehensive and accurate representation of the indoor environment, leading to improved localization performance.

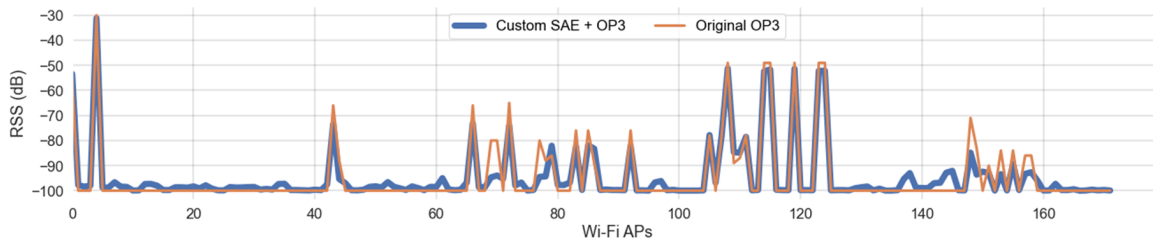

Figure 10: Comparison of Original Fingerprint and Augmented Fingerprint Using Custom SAE on Device: OP3

## 6.3 Comparison with Prior FL and Non-FL Indoor Localization Frameworks

In this section, we present experimental results of comparing *FedHIL* with other state-of-the-art frameworks. We begin by comparing the performance of *FedHIL* with FL and non-FL based frameworks that have been specifically curated for device heterogeneity. The FL frameworks evaluated are *FedHIL*, CRNP [40], FedAeDNN [41], and FedLoc [42], while the non-FL frameworks are ANVIL [33], SHERPA [36], VITAL [37], and WiDeep [39]. Figure 11 shows box plot results from this comparison study. The presented data in the figure is the average error of each framework with all testing client devices across all building paths.

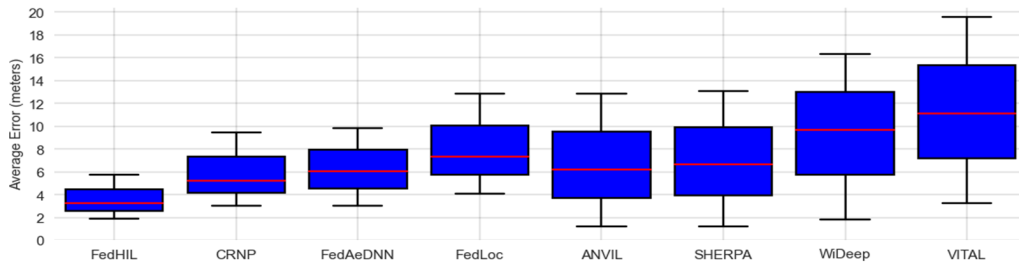

Figure 11: Average localization across all smartphones and floorplans for comparison with FL and non-FL frameworks

It can be observed that the FL frameworks demonstrate superior results compared to the non-FL frameworks in terms of average localization error. This indicates that FL, which leverages multiple devices to train a model collectively, is more effective in addressing device heterogeneity, resulting in more accurate localization. *FedHIL* outperforms all other FL (and non-FL) frameworks, with the lowest average error of 3.24 meters. Specifically, compared to the FL-based indoor localization frameworks CRNP, FedAeDNN, and FedLoc, our *FedHIL* framework is able to achieve 1.62×, 1.86×, and 2.26× lower average localization error, respectively. The non-FL frameworks ANVIL and SHERPA perform better than FedLoc, but still have higher average errors compared to the other FL frameworks. ANVIL and SHERPA use more powerful ML models such as attention networks and KNN with DNN, which explains their better performance. However, the FL frameworks use simpler DNN models and still outperform the complex non-FL frameworks. Analyzing the minimum and maximum errors of each framework, *FedHIL* has the lowest minimum and maximum error among all the frameworks. CRNP, FedAeDNN, and FedLoc have similar minimum and maximum errors. Among the non-FL frameworks, ANVIL and SHERPA have the lowest minimum errors, but their maximum errors are higher than all FL frameworks excluding FedLoc. We observe that WiDeep and VITAL show higher errors compared to other frameworks as these models use GPC (WiDeep) and vision transformers (VITAL) to perform their computation. These models are larger, more complex, and require huge amounts of data for effective performance, indicating the need for a larger database of heterogeneous data to effectively utilize the potential of their underlying models.

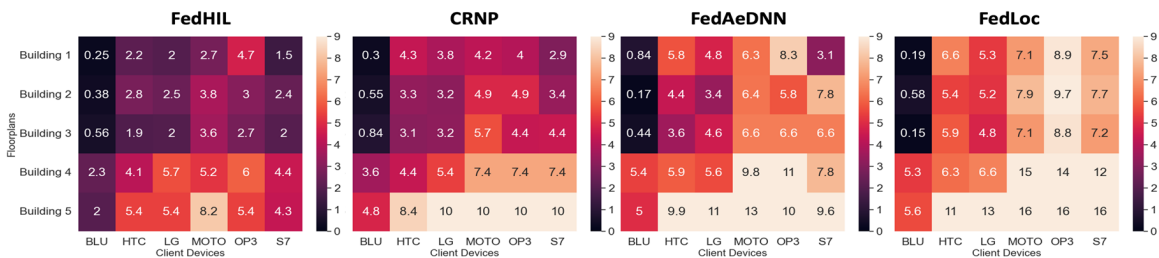

(a): Device-specific performance across all building floorplans and FL frameworks with training device: BLU

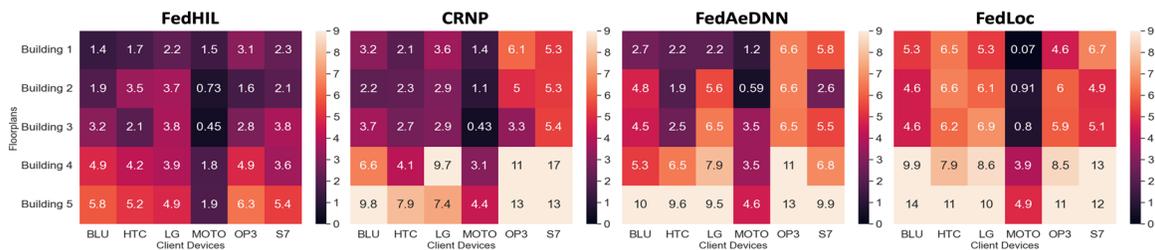

(b): Device-specific performance across all building floorplans and FL frameworks with training device: MOTO

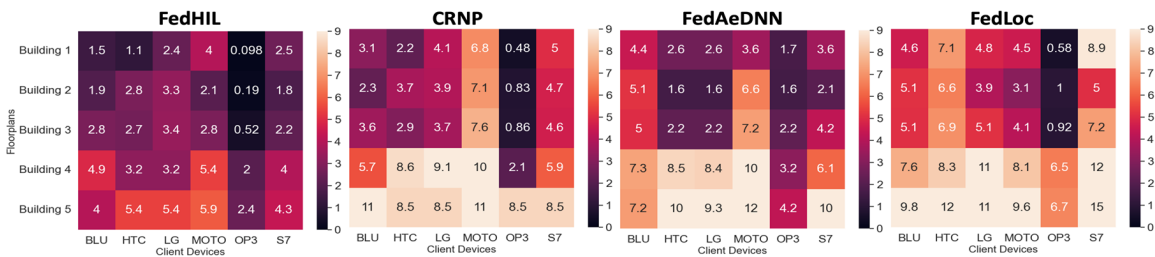

(c): Device-specific performance across all building floorplans and FL frameworks with training device: OP3

Figure 12: Device- specific performance across all building floorplans and FL frameworks for different training devices.

To better understand the performance of the FL frameworks, we present a detailed heatmap view that shows the average error values for each device across the five buildings. In Figure 12, each box indicates the average error for each building floorplan and smartphone. In this analysis, we present results for three training devices. In Figure 12(a), we trained all the frameworks on BLU, Figure 12(b) on MOTO, and Figure 12(c) on OP3. We omit the results from training on the remaining three devices for brevity (trends were similar to those observed in Figure 12). From Figure 12, it can be seen that FedLoc performs well on floorplans with less noise (Building 1, 2, and 3) but fails to provide accurate results on high noise floorplans (Building 4 and 5), even for the same training device that the model was trained on. FedAeDNN has better performance than FedLoc, but still fails to provide accurate results in Building 4 and Building 5. However, it significantly reduces errors from FedLoc, indicating greater resilience towards heterogeneity. CNRP outperforms FedLoc and FedAeDNN since the SAE used in it is better suited for heterogeneity. The FedAvg aggregation approach, used in CNRP and FedAeDNN, performs better than FedLoc's FedSGD aggregation approach. The weighted averaging of FedAvg prevents overfitting, while FedSGD overfits due to gradient aggregation, making it more appropriate for less noisy environments. *FedHIL* exhibits the lowest localization errors for each device compared to other FL-based frameworks. It provides accurate results even in floorplans with high noise (Building 4 and 5), effectively mitigating bottlenecks observed in other frameworks (e.g., in Figure 12(b) see: Building 1-OP3/S7, Building 2-OP3/S7, Building 3-S7, Building 4-LG/OP3/S7, and Building 5-BLU/HTC/LG/MOTO/OP3/S7).

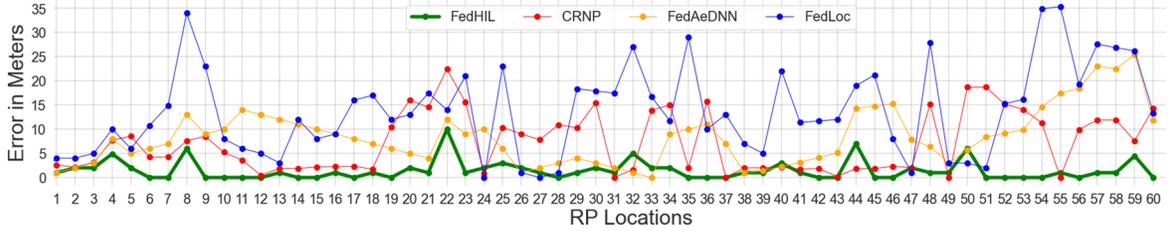

Figure 13: Susceptibility to noise across FL frameworks with training device : MOTO, testing device OP3 in Building 5.

Figure 13 depicts the susceptibility of various FL-based frameworks to noisy data collected in the online phase at each RP, including *FedHIL*, in the high-noise environment of Building 5 with MOTO as the training device and OP3 as the testing device. The noisy environment in Building 5 led to many instances of introducing noisy or outlier fingerprints, leading to the training of the ML model with noisy data, which can degrade the performance of FL-based frameworks. Our *FedHIL* framework and its selective aggregation technique, designed to prevent the underlying SNN model from degrading due to noise, outperforms other FL frameworks. The combination of the proposed SAE and the novel aggregation technique reduces the error even on highly noisy RPs, demonstrating the superior performance of *FedHIL* in noisy environments.

### 6.4 Skew and Scalability Analysis

In this section, we show results of further experiments to explore *FedHIL*'s resilience to skewness and scalability. Skewness refers to uneven distribution of device types in the FL system. Scalability refers to the FL system's ability to handle an increase in clients and data.

#### 6.4.1 Skewness analysis

A skewness analysis was conducted using six different users, each with a smartphone (client). The client device combinations were split into five cases, with each case consisting of a different combination of device types. All the cases were initially trained on the MOTO device and evaluated on all the building floorplans. In Figure 14, we plot the average localization error per case per floorplan with all device combinations in the case split. The device combinations in each of the cases are as follows : Skew Case 1 contains two unique devices (5 users with BLU and 1 with S7), Skew Case 2 contains three unique devices (4 users with BLU, 1 with OP3, and 1 with S7), Skew Case 3 contains four unique devices (3 users with BLU, 1 with MOTO, 1 with OP3, and 1 with S7), Skew Case 4 contains five unique devices (2 users with BLU, 1 with LG, 1 with MOTO, 1 with OP3, and 1 with S7), and Skew Case 5 contains six unique devices (each user has a different device: BLU, MOTO, LG, OP3, S7, and HTC).

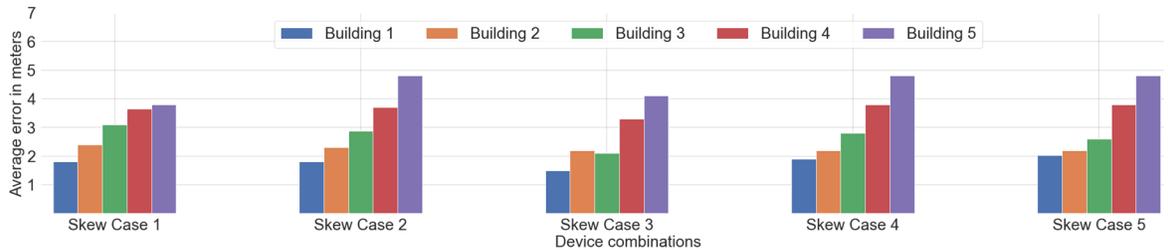

Figure 14: Skewness analysis on different floorplans and devices with different device mixes among clients.

The results in Figure 14 indicate that the *FedHIL* framework is fairly resilient to device skewness. Even in the presence of skewness, the model performance remains relatively stable across scenarios (comparing for the same building floorplan). In Skew Case 1, we observe lower localization errors in all buildings. This is because in the online training phase, the data is weighted more for a particular device, and hence the model learns to generalize better, even when there is one other heterogeneous device. In Skew Case 2, we observe similar results as that of Skew Case 1, but with a slight error increase for Building 5. This is due to the high-noise environment in that building. The Skew Case 3 shows similar results as for Skew Case 1 and Skew Case 2. The Skew Case 4 and Skew Case 5 scenarios show similar results even in high noise environments but show slightly higher errors compared to Skew Case 3, due to the added diversity in the participating client devices. Although Skew Cases 4 and 5 contain a greater number of heterogeneous client devices, they show similar performance as in Skew Case 2. From the results, we can conclude that the *FedHIL* framework shows resilience to device skewness, making it practical for deployment in real-world scenarios where it can provide good accuracy even when the device heterogeneity mix varies among clients.

*6.4.2 Scalability analysis*

This scalability analysis explored the performance of *FedHIL* as the number of clients was increased from 6 to 12 and 18. The 6 Client case had six users with unique smartphone devices, the 12 Client case had twelve devices with each of the six unique smartphone devices used by two out of the 12 users, and the 18 Client case had eighteen devices with each of the six unique smartphone devices used by 3 out of the 18 users. In all scenarios, we set the training device to be MOTO, and the trained model was tested on all participating devices in all available building floorplans.

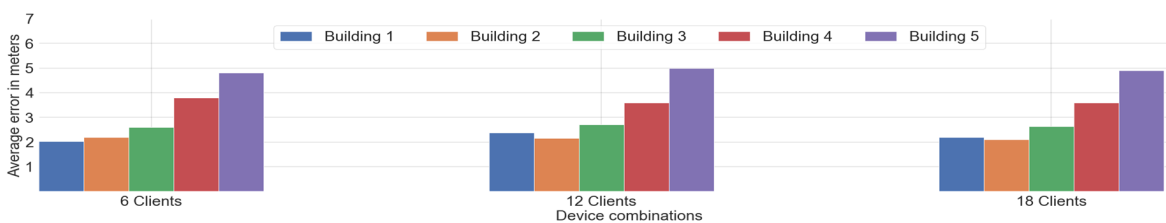

Figure 15: Scalability analysis on different floorplans and varying devices combinations split.

Figure 15 shows the average localization error per scenario per floorplan. Our findings demonstrate similar localization results across all client devices, with little to no variations across the scenarios. This indicates that the *FedHIL* framework is scalable, and its performance is stable, even in the presence of scalability of the participating clients. This is an important observation which suggest that the *FedHIL* framework has great potential for scalability in indoor localization using Wi-Fi RSS fingerprinting.

## 7 CONCLUSION

In this paper, we presented *FedHIL*, a new FL-based framework that improves indoor localization accuracy while preserving user privacy across diverse environments. *FedHIL* employs a custom SAE, a lightweight neural network, and a novel aggregation strategy in a distributed learning setting, thereby diversifying the training data and improving the model's accuracy. To address privacy concerns and robustness, *FedHIL* employs a domain-specific selective weight adjustment approach. We performed extensive analysis with the proposed framework, testing it for heterogeneity, susceptibility to noise, scalability, and skewness, which are common situations faced in real-world distributed learning deployments. Experimental evaluations in real-world settings showed that *FedHIL* outperformed state-of-the-art FL- and non-FL based frameworks, achieving 1.62× better localization accuracy on average than the best performing FL-based indoor localization framework from prior work. *FedHIL* thus represents a promising practical framework for simultaneously addressing the many challenges related to heterogeneity, noise-resilience, data privacy, and accuracy in the indoor localization problem domain.

**ACKNOWLEDGEMENTS**

This research is supported in part by the National Science Foundation grant CNS-2132385.

# REFERENCES

[1] Langlois, C., Tiku, S., Pasricha, S., 2017. Indoor localization with smartphones, IEEE CEM.

[2] Sadowski, S., Spachos, P., 2018. RSSI-Based Indoor Localization With the Internet of Things, IEEE Access.

[3] Lymberopoulos, D., Liu, J., 2017. The Microsoft indoor localization competition: Experiences and lessons learned, IEEE Signal.

[4] Raza, A., Lolic, L., Akhter, S., Liut, M., 2021. Comparing and Evaluating Indoor Positioning Techniques, IEEE IPIN.

[5] Piao, S., Ba, Z., Su, L., 2019. Automating csi measurement with uavs: from problem formulation to energy-optimal solution, IEEE INFOCOM.

[6] Li, Y., He, Z., Gao, Z., Zhuang, Y., Sheimy, C., 2018. Toward robust crowdsourcing-based localization: A fingerprinting accuracy indicator enhanced wireless/magnetic/inertial integration approach, IEEE IoT.

[7] Zhang, P., Zhao, Q., Li, Y., Ni, X., Zhuang, Y., 2015. Collaborative wifi fingerprinting using sensor-based navigation on smartphones, IEEE Sensors.

[8] Xiao, J., Yi, Y., Wang, L., Li, H., Zhou, Z., 2014. Nomloc: Calibration-free indoor localization with nomadic access points, IEEE ICDCS.

[9] Bose, A., Chuan, F., 2007. A practical path loss model for indoor wifi positioning enhancement, IEEE ICICS.

[10] Basiri, A., Lohan, E., Moore, T., Winstanley, A., 2017. Indoor location-based services challenges, requirements and usability of current solutions, Computer Science Review.

[11] Shixiong, X., Liu, Y., Yuan, G., Zhu, M., Wang, Z., 2017. Indoor Fingerprint Positioning Based on Wi-Fi: An Overview, ISPRS.

[12] 2022. Target rolls out bluetooth beacon technology in stores to power new indoor maps in its app, [Online] http://tcrn.ch/2fbIM0P.

[13] Shang, S., Wang, L., 2022. Overview of WiFi fingerprinting-based indoor positioning, IET Communications.

[14] Singh, N., Choe, S., Punmiya, R., 2021. Machine Learning Based Indoor Localization Using Wi-Fi RSSI Fingerprints: An Overview, IEEE Access.

[15] Eberechukwu, P., Park, H., Laoudias, C., 2022. DNN-based Indoor Fingerprinting Localization with WiFi FTM, IEEE MDM.

[16] Liu, G., Qian, Z., Wang, X., 2020. An Indoor WLAN Location Algorithm Based on Fingerprint Database Processing, IJPRAI.

[17] Luo, M., Zheng, J., Sun, W., Zhang, X., 2021. WiFi-based Indoor Localization Using Clustering and Fusion Fingerprint, IEEE CCC.

[18] Nagia, N., Rahman, M., Valaee, S., 2021. Federated Learning for WiFi Fingerprinting, ICC 2022 - IEEE CCC.

[19] Tasbaz, O., Moghtadaiee, V., 2022. Zone-Based Federated Learning in Indoor Positioning, ICCKE.

[20] Jeong, M., Choi, S., 2023. A tutorial on Federated Learning methodology for indoor localization with non-IID fingerprint databases, ICT express.

[21] Sadowski, S., Spachos, P., 2022. Memoryless Techniques and Wireless Technologies for Indoor Localization With the IoT, IEEE IoT.

[22] Mabunga, Z., Cruz, J., 2021. Utilization of Different Wireless Technologies' RSSI for Indoor Environment Classification Using SVM, IEEE.

[23] Lee, L., Kim, J., Moon, N., 2019. Random Forest and WiFi fingerprint-based indoor location recognition system using smart watch, HCCIS.

[24] Zong, B., Zong, Z., Huang, B., Baker, T., 2021. A DNN-based WiFi-RSSI Indoor Localization Method in IoT, LNICST.

[25] Zhao, B., Zhu, D., Xi, T., Jia, C., Jiang, S., Wang, S., 2019. Convolutional neural network and dual-factor enhanced variational Bayes adaptive Kalman filter based indoor localization with Wi-Fi, Computer Networks.

[26] Li, A., Fu, J., Shen, H., Sun, S., 2021. A Cluster-Principal-Component-Analysis-Based Indoor Positioning Algorithm, IEEE IoT.

[27] Pinto, B., Barreto, R., Souto, E., Oliveira, H., 2021. Robust RSSI-Based Indoor Positioning System Using K-Means Clustering and Bayesian Estimation, IEEE Sensors.

[28] Yang, J., Zou, H., Jiang, H., Xie, L., 2018. Device-free Occupant Activity Sensing using WiFi-enabled IoT Devices for Smart Homes, IEEE IoT.

[29] Gokalp, O., Lee, J., Burghal, D., Molisch, A., 2022. Simple and Effective Augmentation Methods for CSI Based Indoor Localization, arXiv:2211.10790v1.

[30] Tang, Z., Li, S., Kim, K., Smith, J., 2022. Multi-Output Gaussian Process-Based Data Augmentation for Multi-Building and Multi-Floor Indoor Localization, IEEE ICC.

[31] Gufran, D., Tiku, S., Pasricha, S., 2023. SANGRIA: Stacked Autoencoder Neural Networks with Gradient Boosting for Indoor Localization, IEEE ESL

[32] Dong, Y., Arslan, T., Yang, Y., 2022. A WiFi Fingerprint Augmentation Method for 3-D Crowdsourced Indoor Positioning Systems, IEEE IPIN.

[33] Chen, X., Li, H., Zhou, C., Liu, X., Wu, D., 2022. Fidora: Robust WiFi-Based Indoor Localization via Unsupervised Domain Adaptation, IEEE IoT.

[34] Tiku, S., Gufran, D., Pasricha, S., 2022. Multi-Head Attention Neural Network for Smartphone Invariant Indoor Localization, IEEE IPIN.

[35] Tiku, S., Gufran, D., Pasricha, S., 2023. Smartphone Invariant Indoor Localization Using Multi-head Attention Neural Network, Machine Learning for Indoor Localization and Navigation, Springer

[36] Tiku, S., Pasricha, S., Notaros, B., 2019. SHERPA: A Lightweight Smartphone Heterogeneity Resilient Portable Indoor Localization Framework, IEEE ICESS.

[37] Gufran, D., Tiku, S., Pasricha, S., 2023. VITAL: Vision Transformer Neural Networks for Accurate Smartphone Heterogeneity Resilient Indoor Localization, IEEE DAC.

[38] Gufran, D., Tiku, S., Pasricha, S., 2023. Heterogeneous Device Resilient Indoor Localization Using Vision Transformer Neural Networks, Machine Learning for Indoor Localization and Navigation, Springer

[39] Abbas, M., Elhamshary, M., 2019. WiDeep: WiFi-based Accurate and Robust Indoor Localization System using Deep Learning, IEEE PerCom.

[40] Li, W., Zhang, C., Tanaka, Y., 2020. Pseudo Label-Driven Federated Learning-Based Decentralized Indoor Localization via Mobile Crowdsourcing, IEEE Sensors.

[41] Gao, B., Yang, F., Cui, N., Xiong, K., Lu, Y., 2023. A Federated Learning Framework for Fingerprinting-Based Indoor Localization in Multibuilding and Multifloor Environments, IEEE IoT.

[42] Cheng, X., Ma, C., Li, J., Song, H., Shu, F., 2022. Federated Learning-Based Localization With Heterogeneous Fingerprint Database, IEEE WCL.

[43] Brendan, H., Moore, E., Ramage, D., 2016. Communication-Efficient Learning of Deep Networks from Decentralized Data, arXiv.1602.05629.

[44] Konečný, J., McMahan, H., Yu, H., Richtárik, P., 2016. Federated Learning: Strategies for Improving Communication Efficiency, arXiv.1610.05492.

[45] Zafari, F., Gkelias, A., Leung, K., 2019. A Survey of Indoor Localization Systems and Technologies, IEEE CST.

[46] Han, S., Pool, J., Tran, J., Dally, W., 2016. Learning both Weights and Connections for Efficient Neural Networks, NIPS.